# IMPACT OF LIDAR VISUALISATIONS ON SEMANTIC SEGMENTATION OF ARCHAEOLOGICAL OBJECTS

*Raveerat Jaturapitpornchai[1*], Giulio Poggi[1*], Gregory Sech[1], Žiga Kokalj[2], Marco Fiorucci[1], Arianna Traviglia[1]*

[1]Center for Cultural Heritage Technology, Istituto Italiano di Tecnologia, Venice, Italy
[2]Research Centre of the Slovenian Academy of Sciences and Arts, Ljubljana, Slovenia

## ABSTRACT

Deep learning methods in LiDAR-based archaeological research often leverage visualisation techniques derived from Digital Elevation Models to enhance characteristics of archaeological objects present in the images. This paper investigates the impact of visualisations on deep learning performance through a comprehensive testing framework. The study involves the use of eight semantic segmentation models to evaluate seven diverse visualisations across two study areas, encompassing five archaeological classes. Experimental results reveal that the choice of appropriate visualisations can influence performance by up to 8%. Yet, pinpointing one visualisation that outperforms the others in segmenting all archaeological classes proves challenging. The observed performance variation, reaching up to 25% across different model configurations, underscores the importance of thoughtfully selecting model configurations and LiDAR visualisations for successfully segmenting archaeological objects.

## 1. INTRODUCTION

Archaeological prospections based on airborne LiDAR (ALS) data usually rely on the visual inspection of point cloud-interpolated Digital Elevation Models (DEM) [1] - [3]. To enhance the visibility of subtle topographic alterations characteristic of archaeological objects, images are typically processed through a comprehensive set of algorithms, convolutional filters, and image-blending methods, often termed Visualisation Techniques (VTs) [4].

In the last five years, the visual analysis of DEM has been increasingly assisted by the application of Deep Learning (DL) methods, significantly reducing the time needed for processing extensive territorial data [5], [6]. By relying on the good practices established for visual inspection, VTs are generally employed in training DL models. While the use of VTs for visual assessments has been widely acknowledged by the scientific community, their impact on DL performance is often overlooked. Only a handful of studies have deliberately attempted to assess the significance of incorporating VTs in DL models [7], [8]. To our knowledge, Guyot et al. [8] are the sole contributors who have ranked VTs based on the performance achieved on semantic segmentation tasks, rather than object detection. Their experiments, however, lack comprehensiveness, hindering a general extrapolation of results. This limitation stems from the use of a limited archaeological dataset and the unique architecture of the DL model. Advancing beyond the state-of-the-art, this work establishes a comprehensive testing framework aiming at facilitating the selection of appropriate DEM-derived products (VTs) for enhancing the performance of semantic segmentation models in archaeological applications. Specifically, this study assesses the influence of diverse DL configurations and visualisations in identifying five distinct archaeological object classes. The contribution extends existing research in three crucial aspects:

- *broader archaeological context*: considering five classes of archaeological objects from two geographically separated ALS datasets.
- *diverse semantic segmentation models:* deploying and evaluating eight distinct semantic segmentation models characterised by various architectures, encoders, and initialisation procedures.
- *innovative visualisation techniques*: introducing three novel VTs designed to explore DL capabilities in identifying archaeological objects.

## 2. DATASETS

This research utilises two distinct ALS datasets, encompassing diverse territories with varying geological, topographic, climatic, vegetational, and archaeological attributes: the Chactun dataset and the Veluwe dataset. Each dataset comprises tiles (256 pixels per size) generated from a LiDAR-derived DEM featuring a ground sample distance of 0.5 m. For each tile, archaeological objects were manually labelled resulting in a raster mask. The Chactun dataset [9] includes three archaeological classes: aguadas (artificial water reservoirs, 166 tiles), buildings and platforms (3532 and 2472 tiles, respectively) from the Mayan civilisation. Spanning the central region of Yucatan, Mexico, this dataset consists of 3568 tiles. The Veluwe dataset [10], from the Netherlands, comprises 1314 tiles with two classes: barrows (round earthen burial mounds, 998 tiles) and charcoal kilns (circular shallow ditch with a central platform for charcoal production, 328 tiles).

---
[*] Equal contribution

## 3. VISUALISATIONS

A total of 7 VTs were computed using the Relief Visualization Toolbox [4] (fig.1). DEM-c is a stretched version of the DEM, computed by cutting 1% at both tails of the distribution for each tile. SLRM is a trend-removal filter that reduces the impact of large-scale features (e.g. terrain slopes) on the visibility of small-scale archaeological objects. e2MSTP is an enhanced version of the Multiscale Topographic Position index computed to highlight variations across various scales. Notably, this specific VT attained the highest ranking in the semantic segmentation task conducted by Guyot et al. in their work [8]. The VAT is a composite image that stacks Slope, Openness and Sky-View Factor [4], [11] ranked highest in Somrak et al. study [7] on the object detection task over the Chactun dataset.

This paper presents three novel VTs to investigate the capabilities of DL in detecting archaeological objects. DEM-s, a three-layer stack of a single DEM-c, was created to assess potential variations in performance between a single band image and a three bands image. This is done to evaluate the impact of the model's initialisations pre-trained on RGB benchmark datasets (ref. chapter 4). DSS is a stack of DEM-c, Slope (i.e., the first derivative of DEM) and SLRM. Its purpose is to evaluate whether incorporating additional visualisations to DEM-c can enhance performance. e2MSTP-1B integrates e2MSTP into a single-band image, allowing us to evaluate the potential loss of information when transitioning from a three-band image to a single band.

## 4. METHODOLOGY

This study aims to provide a nuanced understanding of the performance of supervised semantic segmentation architectures across a spectrum of seven visualisations. Two renowned networks for semantic segmentation, specifically U-Net [12] and DeepLabV3+ [13], were chosen. The decision to employ U-Net in archaeological research was driven by its proficiency in handling scenarios with limited data. This is attributed to its distinctive architecture, characterised by a contracting path, bottleneck layer, and expansive path, which excels in capturing intricate contextual information while preserving fine details. The incorporation of skip connections further enhances accuracy by integrating low-level features.

On the other hand, DeepLabV3+ was purposefully chosen for its exceptional accuracy in multi-scale segmentation. This sophisticated technique excels in meticulously analysing images at different scales, guaranteeing the thorough capture of objects with various sizes and shapes, qualities frequently encountered in archaeological imagery. These two networks were modified by changing their backbone feature extractors to ResNet [14] and EfficientNet [15], aiming to assess the capability of each encoder in extracting features of archaeological objects. In addition to the variation of the network architectures, the effect of weight initialisation is also investigated by comparing networks pre-trained with a commonly adopted benchmark dataset (ImageNet) and Kaiming initialisation [16]. In total, this paper compares eight models, as the combination of two networks, two encoders and two initialisations for each LiDAR visualisation.

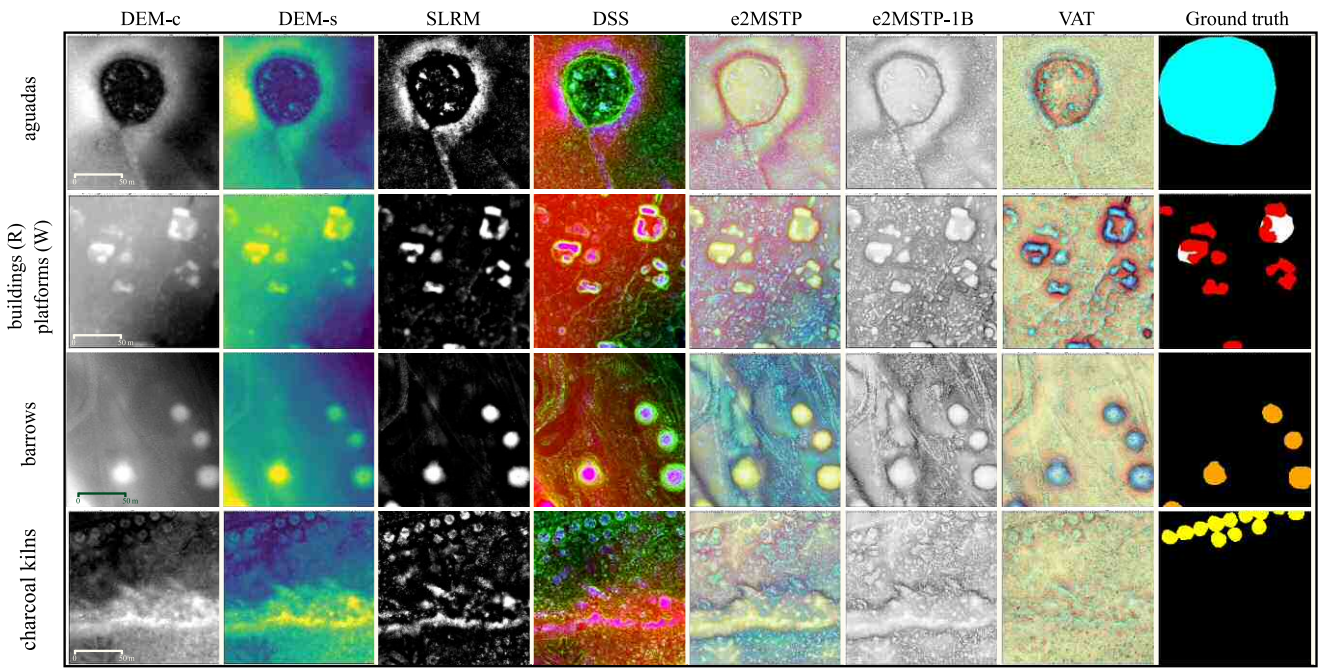

**Fig. 1: Appearance of archaeological classes on different Visualisation Techniques and ground truth mask**

### 4.1. Experimental Setting

The training process spanned 50 epochs, utilising the PyTorch Adam optimiser initialised with a learning rate of 0.001. The choice of Tversky loss, specifically designed for semantic segmentation, was selected for its capacity to handle a good trade-off between precision and recall as outlined by Seyed et al. [17]. Employing a 5-fold cross-validation approach for each model added robustness to the empirical results, mitigating the impact of statistical fluctuations. During the training phase, Kornia's augmentation techniques were leveraged, including vertical flip, horizontal flip, and 45-degree rotation (50% probability). A threshold of 0.5 was employed to determine detected pixels.

The evaluation of segmentation results relied on the Intersection over Union (IoU) per class, precision, and recall. These metrics were deliberately selected to address the class-imbalance issue by minimising the importance of empty tiles when IoU is computed for each class.

## 5. EXPERIMENTAL RESULTS

The comprehensive analysis of the obtained results reveals that buildings is the class with the highest segmentation performance, showcasing the best model with an IoU of 0.66. Following closely are barrows at 0.57, platforms at 0.53, aguadas at 0.47, and charcoal kilns at 0.34 (refer to Fig. 2). The analysis of the best models, depicted in Fig. 2, shows that both DEM-c and DEM-s achieved good performance in all the classes, excelling in the segmentation of buildings and charcoal kilns with IoU of 0.67 and 0.34. Interestingly, their best models alternatively outperform each other in various classes, with no apparent correlation to the characteristics of the archaeological objects. At the same time, these visualisations yield a large difference in performance among the configurations, as shown in Fig. 3. Overall, DEM-s exhibits a slightly superior performance compared to DEM-c attributed to lower variability among configurations. SLRM consistently yields good results across all the classes, although with high variability for various configurations. Notably, it achieves the second highest IoU when detecting buildings (0.65), platforms (0.53), and barrows (0.57). In contrast, models using this visualisation could not accurately classify charcoal kilns.

Models utilising DSS excelled in detecting barrows with the highest IoU of 0.57 and ranked second for aguadas (0.46). This implies that incorporating additional visualisations into DEM stack could potentially enhance the detection of specific features, though not always applicable, as evidenced by lower performance in charcoal kilns and buildings.

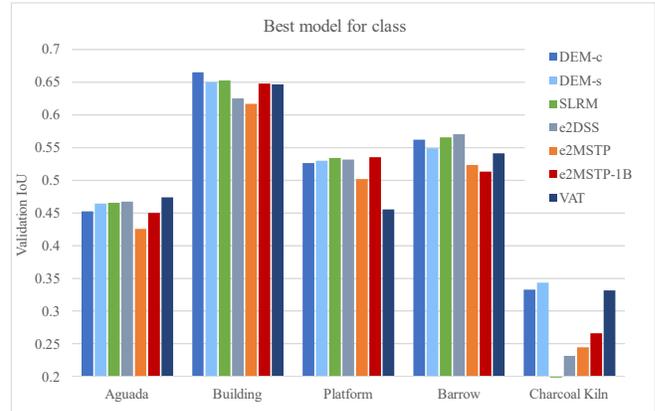

**Fig. 2: Validation IoU of the best model for each visualisation over the five archaeological classes**

Despite e2MSTP overall performance falling below average for every class, it stands out as the most consistent and reliable VT across various configurations, showcasing a narrower range of performance compared to most other VTs. Surprisingly, the e2MSTP-1B generally outperformed e2MSTP, although it displayed a wider variability among the model's configurations. This suggests that flattening the bands into one can be beneficial for extracting more information from the data when selecting suitable configurations.

The VAT performs above average in the other classes, although it exhibits the highest ranking for aguadas with an IoU of 0.47 and exhibits poor performance on platforms (0.45). Despite that, it demonstrates the lowest variability among configurations in the Chactun dataset. Upon analysing the results presented in Fig. 4, it becomes evident that model 5 stands out as the best configuration in both datasets, with models 1, 6 and 7 closely following. Notably, these models consistently exhibit performance across the various visualisations, as underscored by the lower variability compared to the other models.

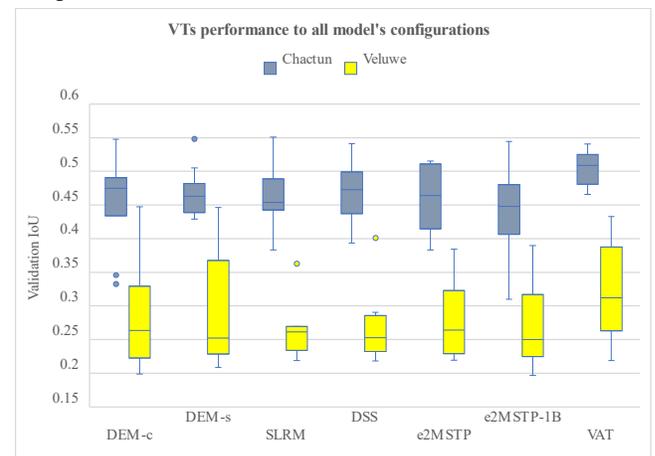

**Fig. 3: Validation IoU of all the models over each VT**

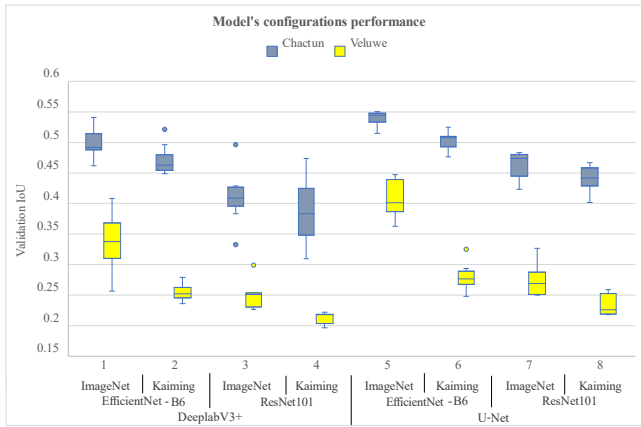

**Fig. 4:** Validation IoU of all the VTs over each model

However, when specifically focusing on the Veluwe dataset alone, variability increases, primarily due to mixed results in detecting charcoal kilns whose low performance is likely due to subtle topographic change and low visibility. Nevertheless, models 5 and 1 consistently outperform other models, while models 2, 3, 4, 6, 7, and 8 exhibit minimal fluctuation when selecting various VTs concerning the validation IoU. Noteworthy is the observation that models 1 and 5 display low fluctuation of performance when opting for different visualisations yet manage to achieve the highest IoU in both datasets. This suggests that, in specific scenarios, the appropriate choice of configuration mitigates variability of performance across the visualisations, emphasising the increased significance of configuration over visualisation selections.

## 6. DISCUSSION

The experimental results underscore the pivotal role of visualisations in DL models, with the potential to influence performance by up to 8%. Despite the absence of a clear dominance among various visualisations, the thoughtful selection of VTs proves crucial in preventing the misclassification of specific classes, as evidenced in the case of charcoal kilns, and enhance segmentation quality. When choosing the evaluation criteria, it is essential to consider not only the results of the best model but also how visualisations affect the variability of results across various model configurations. For example, DEM-c and DEM-s excel in performance but show a high variance potentially impacting the training procedure's reliability.

Opting for more consistent VTs, such as VAT, e2MSTP or SLRM, can be beneficial to avoid unforeseen performance declines, especially with limited configuration choices. Explore architectural nuances is recommended to potentially enhance overall performance. It is advisable considering experiments with DEM-s, one of the novel VTs introduced in this study, for the excellent results achieved in all the classes, with outstanding performance in charcoal kilns segmentation. Comparison between DEM-c and DEM-s revealed that a three-band image has lower performance variability than a single band image. On the other hand, leveraging a single band image, in conjunction with specific configurations, can attain the highest performance, particularly for buildings and barrows. Similar behaviour is observed between three-band e2MSTP and single-band e2MSTP-1B. Contrary to Guyot et al. [8], our findings did not identify e2MSTP as a clear contributor to the top-performing model. Nevertheless, we align with Somrak et al. [7] on the VAT visualisation's good performance, especially in the Chactun dataset, suggesting its suitability for segmentation tasks.

The findings reveal the ability to pinpoint optimal model configurations that excel across all visualisations and classes. Thoughtful choices in architectures, encoders, and initialisations can significantly boost performance, achieving improvements of up to 25%. The configuration employing U-Net, EfficientNet-B6 encoder and initialised with ImageNet consistently emerged as the top performer, frequently achieving the highest IoU on most visualisations. The U-Net architecture outperforms DeeplabV3+, mainly due to the inclusion of skip connections facilitating the detection of complex shapes, particularly in the building class. The primary performance gap stems from encoder selection, favouring EfficientNet-B6 over ResNet101. EfficientNet-B6's superior performance is attributed to its compound scaling method, enabling more effective feature extraction with a comparable number of parameters [15], consistent with prior experiments on these datasets [18].

## 7. CONCLUSION

This study emphasises the key role of selecting LiDAR-derived DEM visualisations to enhance the efficacy of deep learning models for supporting archaeological prospection. Notably, the previously unexplored importance of model configurations is underscored in contrast to various visualisations within the existing literature. The observed synergy between visualisations and model configurations presents promising avenues for advancing future applications and leveraging deep learning in archaeological research.

Our future goal is to deploy this framework across different archaeological classes located in various geographical areas, incorporating additional model configurations to further enrich the significance and impact of our study.

**Acknowledgement**

MF contributed to this study as part of the OPTIMAL project: this project has received funding from the European Union's Horizon 2020 research and innovation programme under grant agreement No 101027956.
The authors thank PhD Wouter Verschoof-van Der Vaart for providing the Veluwe dataset.